\documentclass[10pt,twocolumn,letterpaper]{article}

\usepackage[pagenumbers]{wacv} 

\usepackage{graphicx}
\usepackage{amsmath}
\usepackage{amssymb}
\usepackage{booktabs}

\usepackage{makecell}
\usepackage{multirow}
\usepackage[table]{xcolor}

\usepackage[pagebackref,breaklinks,colorlinks]{hyperref}

\usepackage[capitalize]{cleveref}
\crefname{section}{Sec.}{Secs.}
\Crefname{section}{Section}{Sections}
\Crefname{table}{Table}{Tables}
\crefname{table}{Tab.}{Tabs.}

\def\wacvPaperID{2278} 
\def\confName{WACV}
\def\confYear{2025}

\begin{document}

\title{LTCXNet: Advancing Chest X-Ray Analysis with Solutions for Long-Tailed Multi-Label Classification and Fairness Challenges}

\author{*\textsuperscript{1}Chin-Wei Huang\\
{\tt\small winston78934546@gmail.com}
\and
*\textsuperscript{1}Mu-Yi Shen\\
{\tt\small muyishen2040@gmail.com}
\and
\textsuperscript{1}Kuan-Chang Shih\\
{\tt\small eric12345566@gmail.com}
\and
\textsuperscript{1}Shih-Chih Lin\\
{\tt\small leolin65@gapp.nthu.edu.tw}
\and
**\textsuperscript{2}Chi-Yu Chen\\
{\tt\small altis5526@ntuh.gov.tw}
\and
**\textsuperscript{1}Po-Chih Kuo\\
{\tt\small kuopc@cs.nthu.edu.tw}
\and
\textsuperscript{1}National Tsing Hua University, Taiwan, \textsuperscript{2}National Taiwan University Hospital\\
\small
*Co-first author, **Co-corresponding author
}

\maketitle

\begin{abstract}
Chest X-rays (CXRs) often display various diseases with disparate class frequencies, leading to a long-tailed, multi-label data distribution. In response to this challenge, we explore the Pruned MIMIC-CXR-LT dataset, a curated collection derived from the MIMIC-CXR dataset, specifically designed to represent a long-tailed and multi-label data scenario. We introduce LTCXNet, a novel framework that integrates the ConvNeXt model, ML-Decoder, and strategic data augmentation, further enhanced by an ensemble approach. We demonstrate that LTCXNet improves the performance of CXR interpretation across all classes, especially enhancing detection in rarer classes like `Pneumoperitoneum' and `Pneumomediastinum' by 79\% and 48\%, respectively. Beyond performance metrics, our research extends into evaluating fairness, highlighting that some methods, while improving model accuracy, could inadvertently affect fairness across different demographic groups negatively. This work contributes to advancing the understanding and management of long-tailed, multi-label data distributions in medical imaging, paving the way for more equitable and effective diagnostic tools.
\end{abstract}

\section{Introduction}
\label{sec:intro}

Deep learning Chest X-ray (CXR) models face significant hurdles, including long-tailed distribution and multi-label classification~\cite{cite:PruneCXR-LT,cite:CXR-LT}. Long-tailed distribution, characterized by skewed disease frequency, biases predictive models towards common conditions, undermining the detection of less frequent but critical diseases~\cite{cite:long-tailed_problem2,cite:feature_decoupling,cite:long-tailed_problem}. Accurately identifying these diseases is paramount, particularly when they pose severe health risks~\cite{cite:fatal_rare_problem2,cite:fatal_rare_problem1}.

The complexity of medical image prediction is intensified by the multi-label nature of CXRs~\cite{cite:multilabel_problem1}, where a single image may exhibit multiple diseases. This scenario demands precise disease predictions, necessitating advanced classifiers specifically designed for multi-label tasks~\cite{cite:multilabel_model1,cite:multilabel_model3,cite:ML-Decoder}. These challenges also contribute to fairness concerns in medical imaging classification, emphasizing the need for equitable accuracy across different demographic groups~\cite{cite:fairness_problem1,cite:fairness_problem2}.

Our study is dedicated to CXR image prediction, introducing LTCXNet—a combination of ConvNeXt~\cite{cite:ConvNeXtv1}, ML-Decoder~\cite{cite:ML-Decoder}, data augmentation, and ensemble techniques~\cite{cite:ensemble}—each chosen for its distinct advantage. ConvNeXt's transformer-inspired architecture offers enhanced performance over conventional convolutional neural network (CNN) models, making it an excellent choice for our base architecture. ML-Decoder, an advancement over transformer heads, excels in multi-label classification and reduces computational load. Data augmentation improves both model accuracy and fairness, while ensemble techniques merge insights from various models for more accurate and dependable predictions. We assessed different approaches using the Pruned MIMIC-CXR-LT dataset~\cite{cite:PruneCXR-LT}, focusing on their performance and fairness impact. Our findings demonstrate LTCXNet's superior performance, highlighting its effectiveness in addressing long-tailed and multi-label classification challenges in medical imaging.

\section{Method}
\subsection{Overview}
\begin{figure*}
    \centering
    \includegraphics[width=0.8\linewidth]{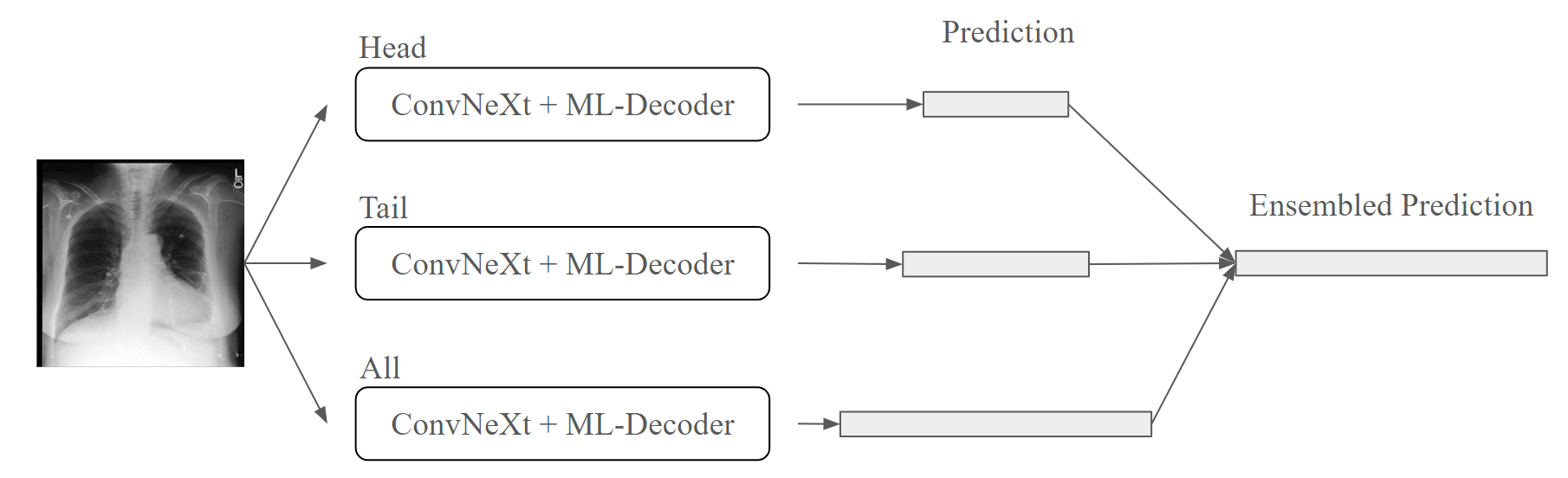}
    \caption{Overview of the proposed method: input image processed by three branches (Head, Tail, All), followed by individual predictions and final ensembled prediction (the length of prediction block indicates the output class number).}
    \label{fig:overview_plot}
\end{figure*}

Fig.~\ref{fig:overview_plot} outlines our model architecture.  We train three distinct models, each focusing on a specific set of labels: `Head', `Tail', and `All'. Each model comprises ConvNeXt, positional encoding \cite{cite:positional_encoding}, and the ML-Decoder. Final predictions for the `Tail' and `Head' classes are averaged with the corresponding class prediction in the `All' model, except for the 'support device' class, which will be further explained in ensemble learning section.

\subsection{Dataset}
We utilize the Pruned MIMIC-CXR-LT~\cite{cite:PruneCXR-LT}, which comprises 257,018 frontal CXRs, each labeled with one of 19 clinical findings. Tailored to address the challenges of long-tailed, multi-label classification of thoracic diseases in CXR images, this dataset addresses the complexity of real-world medical imaging, where a few common findings are followed by many rarer conditions.

\begin{figure*}
    \includegraphics[width=\textwidth]{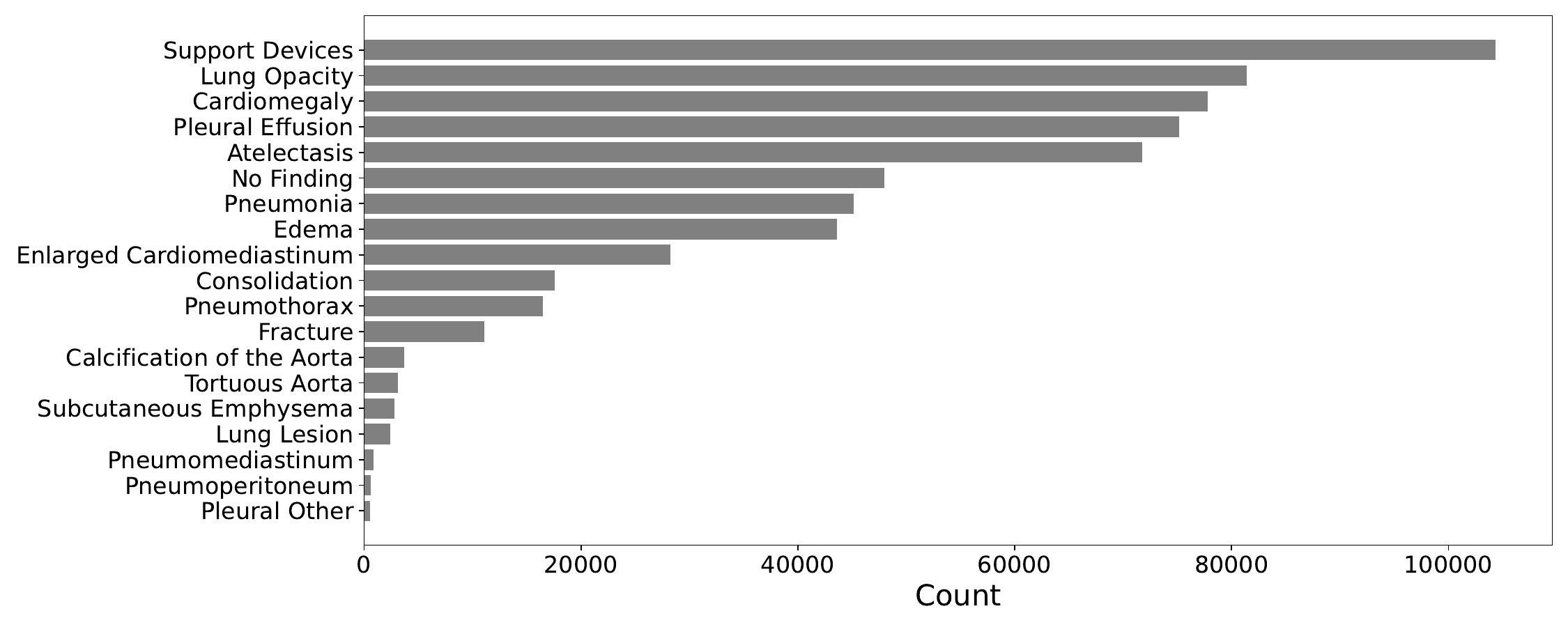}
    \caption{Class distribution in the Pruned MIMIC-CXR-LT dataset.}
    \label{fig:label_count}
\end{figure*}

In our study, the dataset was divided into training, validation, and testing sets, containing 182,380, 20,360, and 54,268 images, respectively. Images undergo resizing to a uniform dimension of $256\times256$ pixels before model input. Fig.~\ref{fig:label_count} illustrates the long-tailed nature of the dataset, with the most common class having 104,364 samples compared to just 553 samples in the least common class, highlighting the substantial disparity in class frequency.

. 

\subsection{ConvNeXt and ML-Decoder}
ConvNeXt~\cite{cite:ConvNeXtv1} blends the robust feature extraction of CNNs with the contextual comprehension of attention models, enhancing both depth and width for comprehensive image analysis. This innovative approach, combined with its proven efficacy in classification, transfer learning, and domain adaptation, making it the chosen backbone for feature extraction. 
In multi-label classification, transformer-decoder architectures prove beneficial for datasets with a limited class range but struggle to scale due to their high computational requirements, which grow quadratically with class size. On the other hand, ML-Decoder~\cite{cite:ML-Decoder}, which also utilized in a similar task~\cite{cite:CheXFusion}, offers a viable alternative by modifying the traditional transformer decoder framework, notably through the removal of self-attention mechanisms to improve efficiency and the adoption of group-decoding to increase scalability independently of class count. Motivated by the reduced computational demands and its demonstrated success, we selected the ML Decoder as our classification head.

\subsection{Augmentation}
Data augmentation can significantly enhance the performance and robustness of machine learning models~\cite{cite:data_aug_survey}. We have explored and assessed a range of traditional data augmentation techniques, providing insights into our selected methods. Our strategies encompass rotation to acquaint models with different perspectives, padding to preserve image consistency, brightness adjustments to emulate various lighting environments, Gaussian blur to introduce controlled blurring mirroring real-world scenarios, contrast manipulation to extract better features, and posterization to reduce tonal levels for noise resilience. 

\subsection{Ensemble Learning}
Ensemble models often achieve higher accuracy by leveraging diverse perspectives on the data, reducing the risk of overfitting and lowering the noise in the dataset~\cite{cite:rakotomamonjy2008bci,cite:rokach2010ensemble}. In our study, `Head' and `Tail' are subsampled from the dataset, while `All' represents the entire dataset following the approach outlined in prior research~\cite{cite:ensemble}. In this division, the `Head' encompasses the nine most prevalent classes, while the `Tail' contains the remaining ten categories and the `Support device' and the `All' comprises every class. 
Note that the `Support device' category appears in both the `Head' and `Tail' due to its prevalence in the dataset. Excluding it would result in the `Tail' having an insufficient number of training samples.

\subsection{Evaluation Metrics}
We employ the mean Average Precision (mAP)~\cite{cite:mAP} and the macro F1 score (mF1) to evaluate model performance. The mAP metric computes the average area under the precision-recall curve for each class, while mF1 determines the average F1 score across all classes. Both metrics treat the performance of each class as equally significant, making them particularly appropriate for datasets with imbalanced class distributions.

To assess fairness of our model, we employ the Equality of Opportunity (EO)~\cite{cite:fairness}, which is essential to ensure that the False Negative Rate (FNR) is consistent across different demographic attributes, a critical factor in avoiding the mis-classification of ill patients as healthy. 

We use $D$ to denote the set defined by demographic attributes. For instance, when considering the demographic attribute of gender, $D$ encompasses the demographic categories male and female. The variables $\hat{Y_i}$ and $Y_i$ denote the predicted and actual labels for the $i$th class, respectively. The FNR for the $i$th class, considering demographic categories $a$, is given by:
\begin{equation}
    FNR(i, a) = p(\hat{Y_i} = 0| D = a, Y_i = 1),
\end{equation}
and EO across $m$ classes, we compute:
\begin{equation}
    EO = \frac{1}{m} \sum_{i=1}^{m} \frac
    {\min_{a \in D} FNR(i, a)}
    {\max_{a \in D} FNR(i, a)}.
\end{equation}
Specifically, we utilize the Youden index to locate the optimal cut-off points on the ROC curve for the FNR. 

\subsection{Implementation Details}
Implemented using PyTorch
, our model employs the ConvNeXt-small~\cite{cite:ConvNeXtv1,cite:rw2019timm}, pre-trained on ImageNet~\cite{cite:imagenet}, with all input images resized to $256\times256$ pixels. The binary cross-entropy loss function is utilized in our task. Optimization is carried out using the Lion optimizer~\cite{cite:lion,cite:lion_github}, with a learning rate set at $6 \times 10^{-6}$ and weight decay at $5 \times 10^{-5}$, complemented by the use of GradScaler 
to enhance training efficiency. Batch size is set to 32. 

\section{Results and Discussion}
\subsection{Performance Evaluation}
\begin{figure*}
  \centering
  \begin{subfigure}{0.48\linewidth}
    \includegraphics[width=1\linewidth]{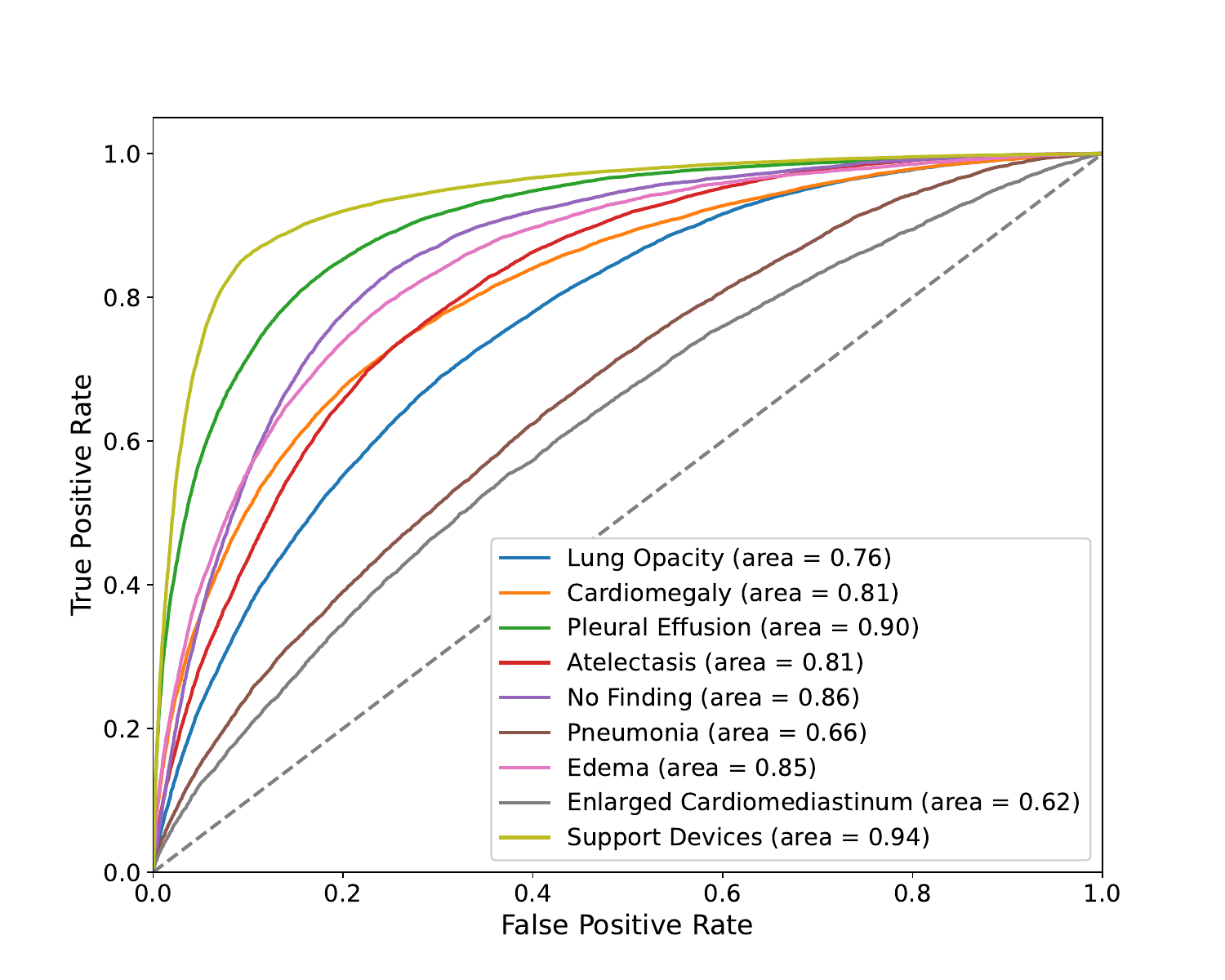}
    \caption{Head}
    \label{fig:short-a}
  \end{subfigure}
  \hfill
  \begin{subfigure}{0.48\linewidth}
    \includegraphics[width=1\linewidth]{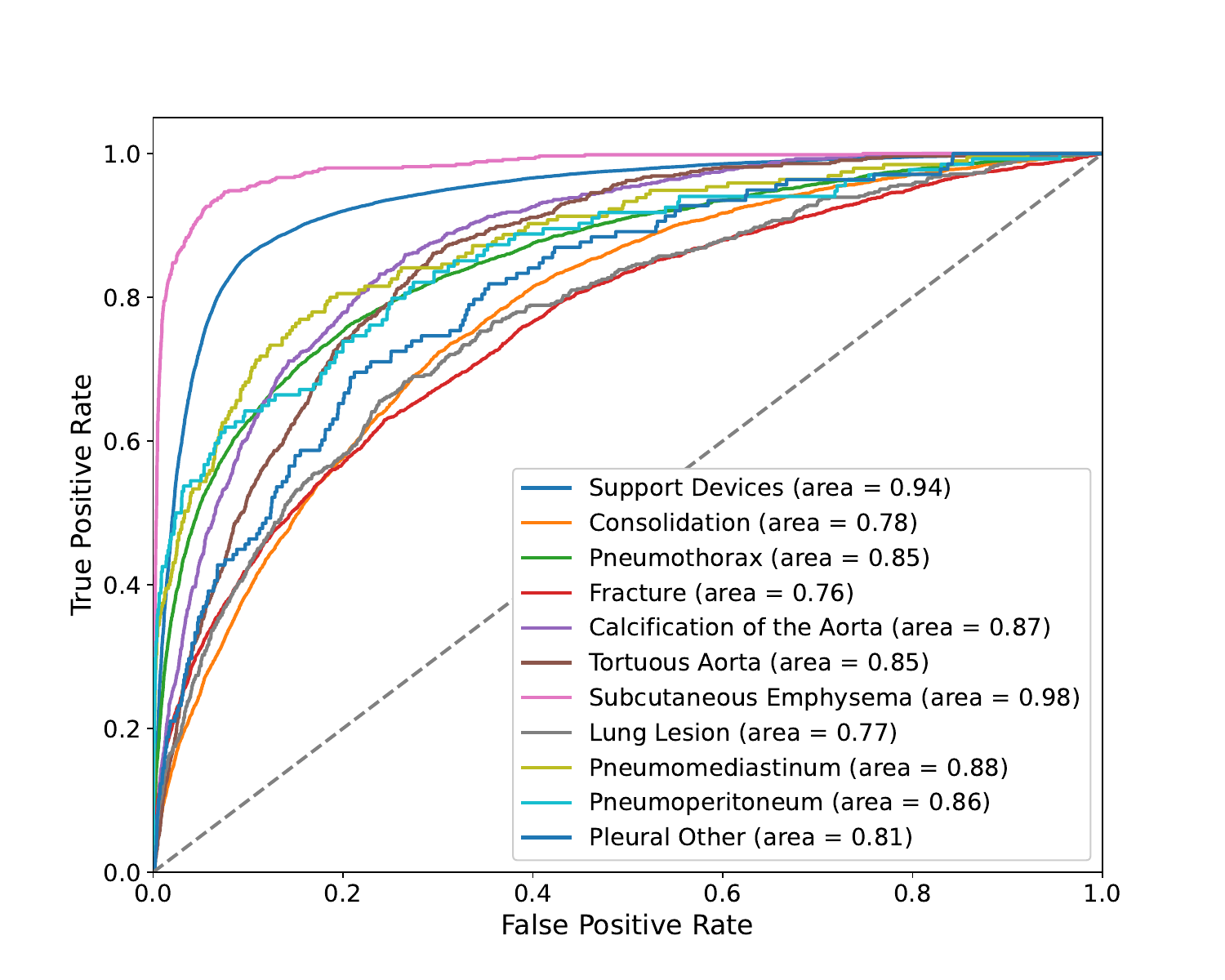}
    \caption{Tail}
    \label{fig:short-b}
  \end{subfigure}
  \caption{ROC Curve and AUC score for 19 classes separated by Head and Tail classes.}
  \label{fig:roc_auc_plot}
\end{figure*}

\begin{figure*}[ht]
    \centering
    \includegraphics[width=\textwidth]{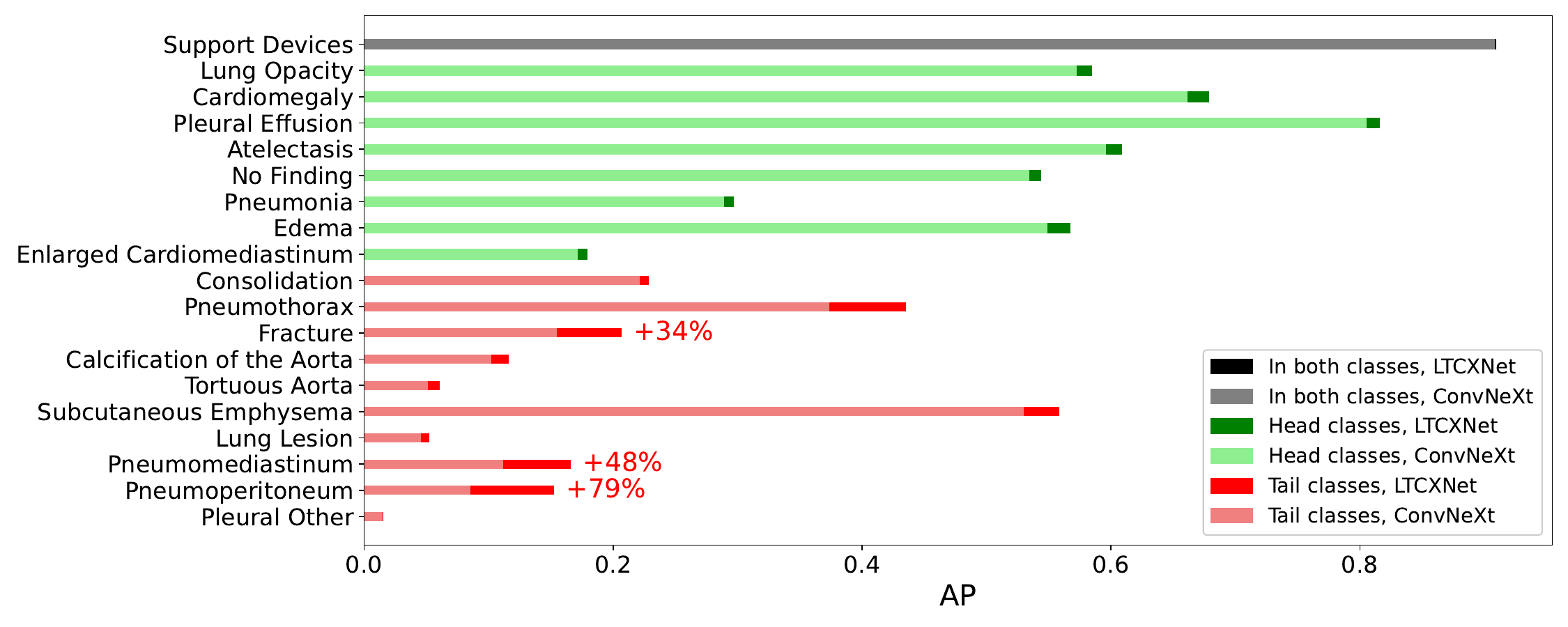}
    \caption{Comparison of LTCXNet and baseline (ConvNeXt): testing AP across disease conditions in CXRs sorted by frequency, highlighting top 3 improvement classes.}
    \label{fig:discussion_APs}
\end{figure*}

In Fig.~\ref{fig:roc_auc_plot}, we present a comparison of the Receiver Operating Characteristic (ROC) Curves and the Area Under the Curve (AUC) score across 19 categories. Due to the nature of AUC calculations, categories with fewer samples may exhibit artificially high AUC values, failing to represent real-world performance accurately~\cite{cite:bad_AUC}. For example, the category of `Subcutaneous Emphysema', which includes a limited number of cases, shows an elevated AUC. This is attributed to the rarity of this condition in the sample set, potentially skewing the perceived accuracy of the model. Consequently, we advocate for using the mAP as a more balanced measure of model efficacy. Fig.~\ref{fig:discussion_APs} organizes diseases along the y-axis by decreasing frequency. Our LTCXNet demonstrates improved performance across all classes compared to the baseline ConvNeXt. Notably, LTCXNet significantly enhances the performance of tail classes. The top three classes with the most improvement are `Pneumoperitoneum', `Pneumomediastinum', and `Fracture', with improvements of 79\%, 48\%, and 34\%, respectively. This underscores LTCXNet's effectiveness in elevating results for these less frequent groups.

\subsection{Backbone Evaluation}
We conducted experiments to compare multiple backbone architectures and determine the most effective model for our task. As illustrated in Table~\ref{table:backbone_table}, the ConvNeXt v1 Small model demonstrated superior performance, achieving the highest validation and testing mAP. This indicates that ConvNeXt v1 Small is the best-suited model for our task.


\begin{table*}[ht]
\centering
\begin{tabular}{cccc}
\toprule
\textbf{Model} & \textbf{GFLOPs} & \textbf{Validation} & \textbf{Test} \\
\midrule
Resnet18 & 2.372 & 0.333 & 0.325 \\
Resnet50 & 5.353 & 0.334 & 0.323 \\
Densenet121 & 3.721 & 0.339 & 0.335 \\
Densenet161 & 10.130 & 0.346 & 0.339 \\
ConvNeXt V1 Small & 11.351 & \textbf{0.357} & \textbf{0.351} \\
ViT & 17.567 & 0.327 & 0.318 \\
Swin Transformer & 15.446 & 0.345 & 0.340 \\
\bottomrule
\end{tabular}
\caption{mAP Performance and GFLOPs of various backbone models.}
\label{table:backbone_table}
\end{table*}

\subsection{Ablation Study}
Table~\ref{table:ablation} presents the outcomes of the ablation study. We observe an incremental trend in mAP as each component is added, while mF1 decreases when ensemble techniques are applied. We choose mAP as the primary evaluation metric since, unlike mF1, which focuses on a single threshold, mAP assesses performance across all thresholds, providing a more comprehensive view. Therefore, we still adopt the ensemble method in LTCXNet. The choice of mAP ensures a more reliable evaluation of the method’s effectiveness, and the results of the ablation study confirm that each component contributes positively to overall model performance.

\begin{table*}[ht]
\centering
\begin{tabular}{|cccc|cc|cc|}
\hline
\multicolumn{4}{|c|}{\textbf{Ablation Settings}} &
  \multicolumn{2}{c|}{\textbf{mAP}} &
  \multicolumn{2}{c|}{\textbf{mF1}} \\ \hline
\multicolumn{1}{|c|}{\textbf{ConvNeXt-S}} &
  \multicolumn{1}{c|}{\textbf{ML-Dec.}} &
  \multicolumn{1}{c|}{\textbf{Data Aug.}} &
  \textbf{Ensemble} &
  \multicolumn{1}{c|}{\textbf{Val.}} &
  \textbf{Test} &
  \multicolumn{1}{c|}{\textbf{Val.}} &
  \textbf{Test} \\ \hline
\multicolumn{1}{|c|}{$\checkmark$} &
  \multicolumn{1}{c|}{$\times$} &
  \multicolumn{1}{c|}{$\times$} &
  $\times$ &
  \multicolumn{1}{c|}{0.357} &
  0.351 &
  \multicolumn{1}{c|}{0.256} &
  0.247 \\ \hline
\multicolumn{1}{|c|}{$\checkmark$} &
  \multicolumn{1}{c|}{$\checkmark$} &
  \multicolumn{1}{c|}{$\times$} &
  $\times$ &
  \multicolumn{1}{c|}{0.366} &
  0.363 &
  \multicolumn{1}{c|}{0.268} &
  0.271 \\ \hline
\multicolumn{1}{|c|}{$\checkmark$} &
  \multicolumn{1}{c|}{$\times$} &
  \multicolumn{1}{c|}{$\checkmark$} &
  $\times$ &
  \multicolumn{1}{c|}{0.370} &
  0.364 &
  \multicolumn{1}{c|}{0.272} &
  0.268 \\ \hline
\multicolumn{1}{|c|}{$\checkmark$} &
  \multicolumn{1}{c|}{$\checkmark$} &
  \multicolumn{1}{c|}{$\checkmark$} &
  $\times$ &
  \multicolumn{1}{c|}{0.376} &
  0.372 &
  \multicolumn{1}{c|}{\textbf{0.313}} &
  \textbf{0.307} \\ \hline
\multicolumn{1}{|c|}{$\checkmark$} &
  \multicolumn{1}{c|}{$\checkmark$} &
  \multicolumn{1}{c|}{$\checkmark$} &
  $\checkmark$ &
  \multicolumn{1}{c|}{\textbf{0.384}} &
  \textbf{0.377} &
  \multicolumn{1}{c|}{0.291} &
  0.287 \\ \hline
\end{tabular}
\caption{Ablation analysis of LTCXNet.}
\label{table:ablation}
\end{table*}

\subsection{Grad-CAM Visualization}
\begin{figure}[t]
  \centering
  \begin{subfigure}{0.32\linewidth}
    \includegraphics[width=0.9\linewidth]{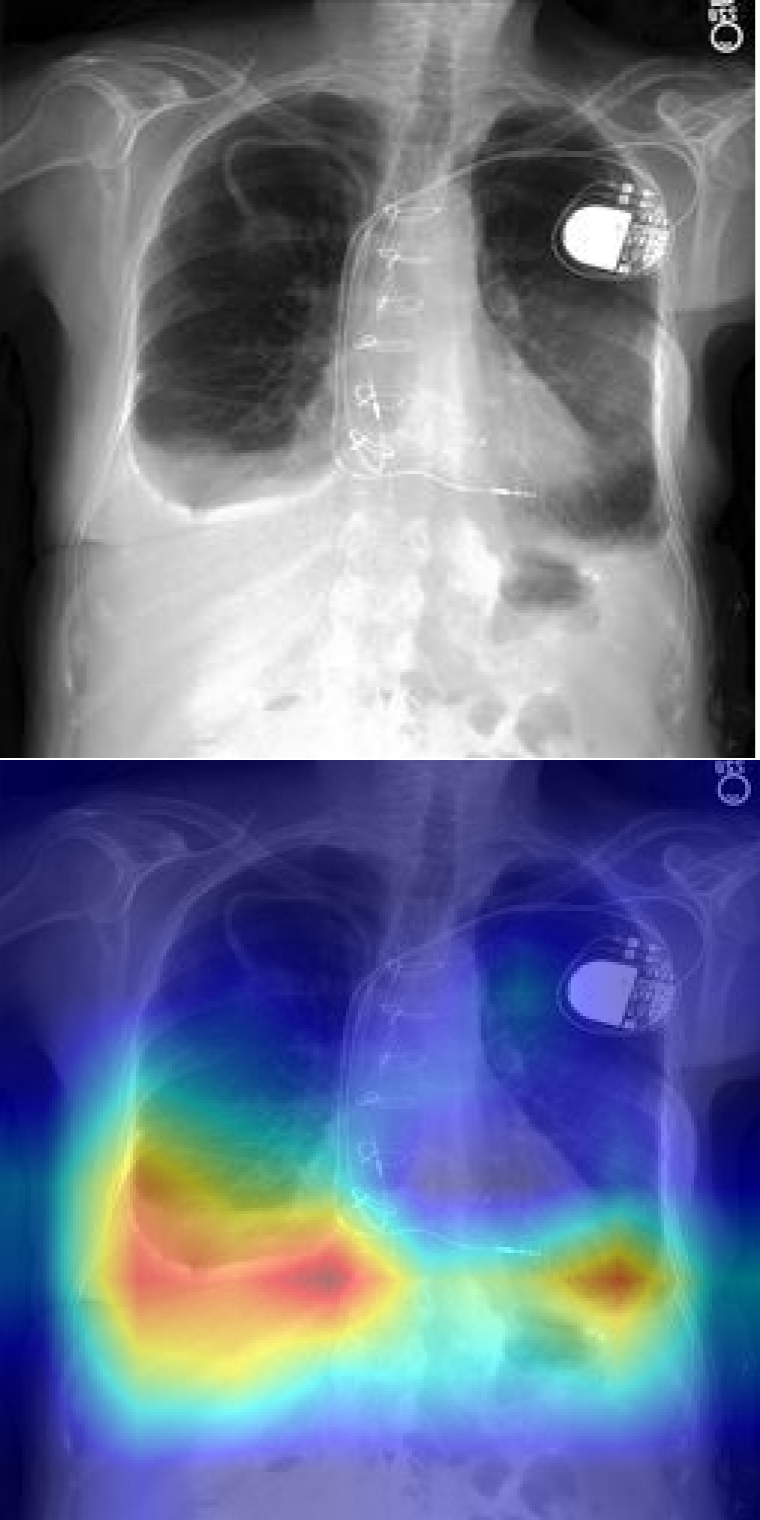}
    \caption{Pleural Effusion}
    \label{fig:Pleural Effusion}
  \end{subfigure}
  \hfill
  \begin{subfigure}{0.32\linewidth}
    \includegraphics[width=0.9\linewidth]{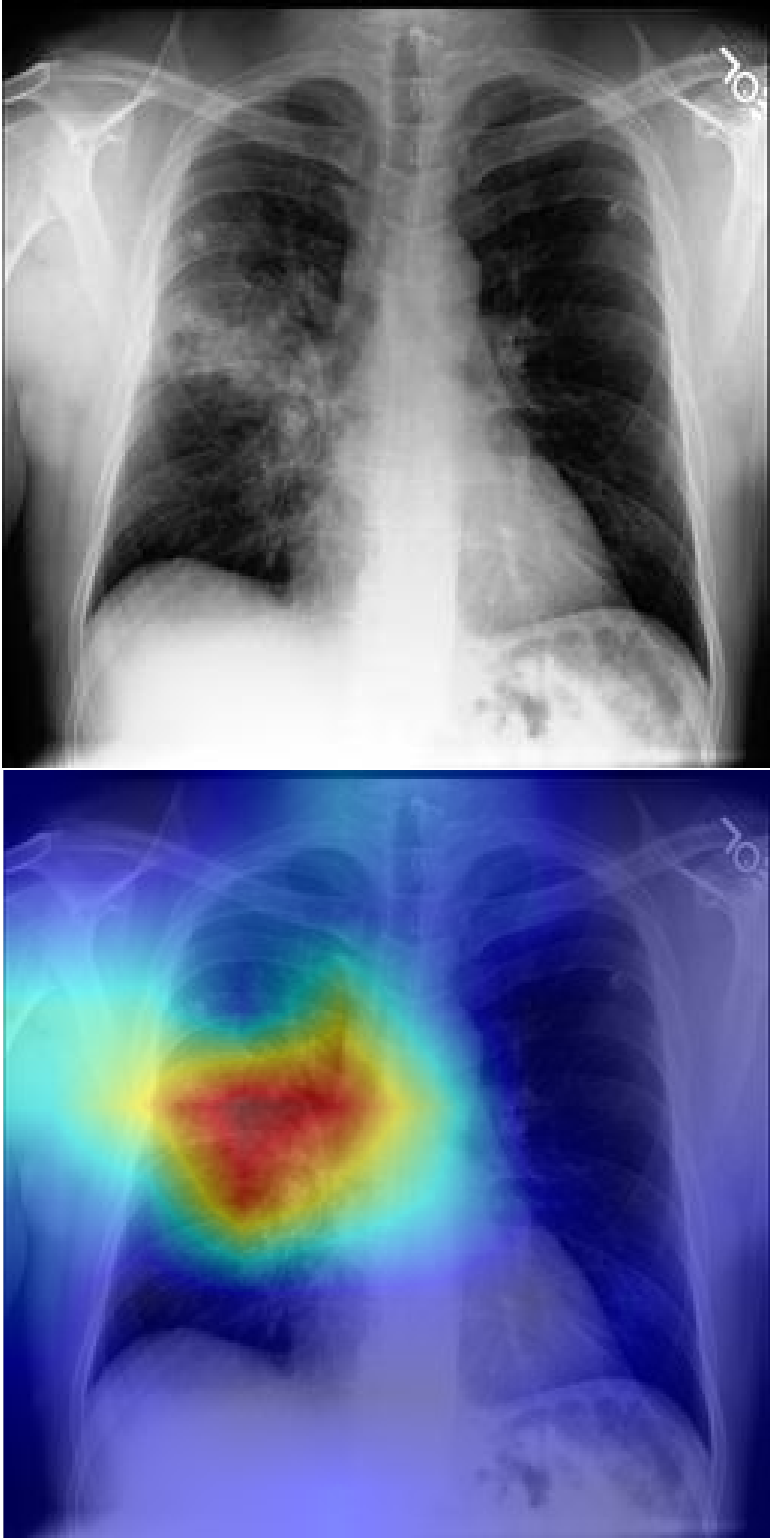}
    \caption{Pneumonia}
    \label{fig:Pneumonia}
  \end{subfigure}
  \hfill
  \begin{subfigure}{0.32\linewidth}
    \includegraphics[width=0.9\linewidth]{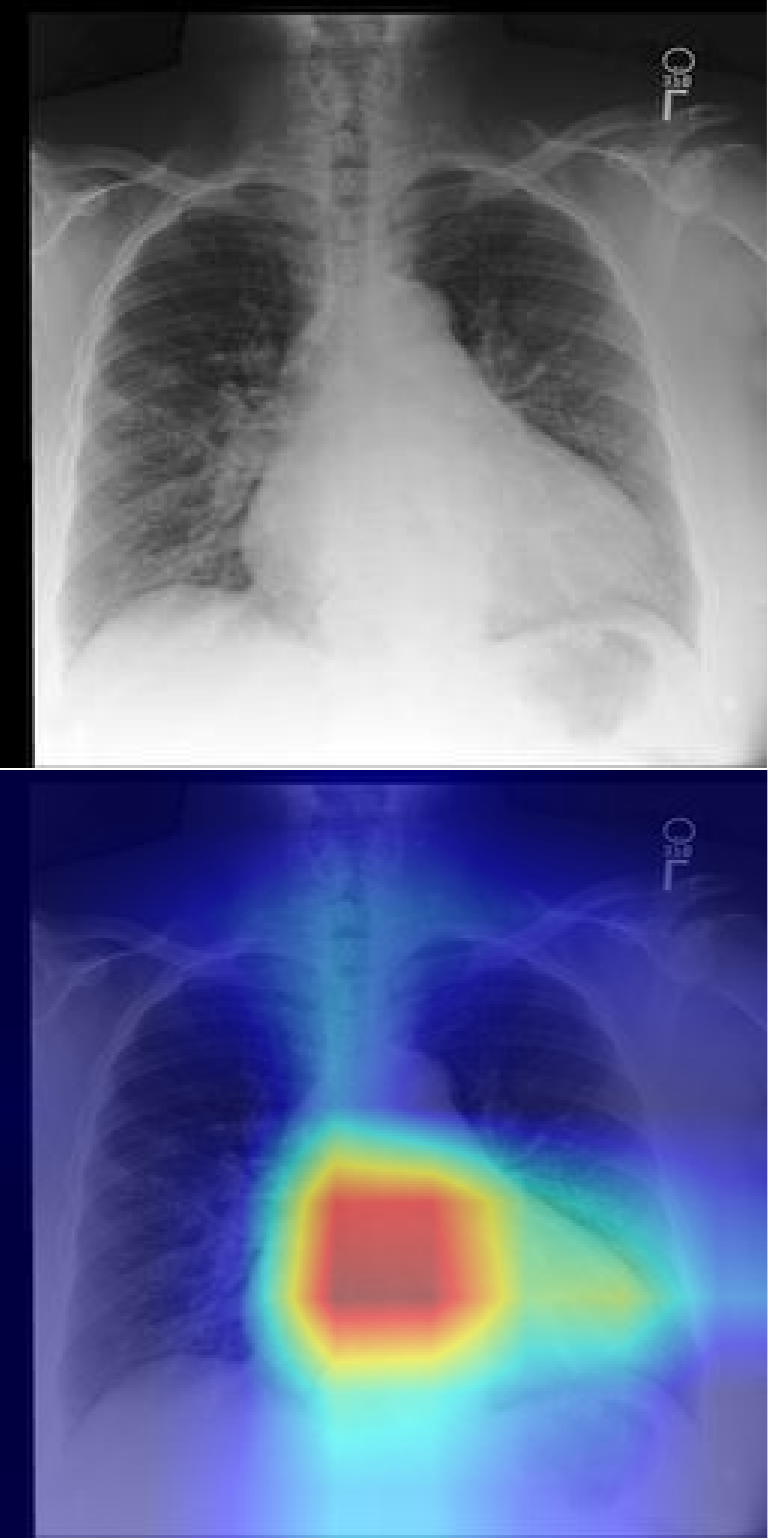}
    \caption{Cardiomegaly}
    \label{fig:Cardiomegaly}
  \end{subfigure}
  \caption{Grad-CAM visualization of (a) Pleural Effusion, (b) Pneumonia, and (c) Cardiomegaly.}
  \label{fig:grad_cam_plot}
\end{figure}

The Grad-CAM~\cite{cite:grad-cam} visualization, depicted in Fig.~\ref{fig:grad_cam_plot}, showcases our model's predictive output for three distinct lung diseases. In this color-coded visualization, red areas indicate the regions the model primarily focuses on, revealing that these highlighted sections of the input CXR image significantly influence the model's predictions. Specifically, Fig.~\ref{fig:Pleural Effusion} illustrates the condition `Pleural Effusion', which typically manifests in the lower lung regions. The Grad-CAM visualization of our model aligns with this clinical presentation, confirming that the focused region corresponds to the actual lesion site. This concurrence underscores the model's practical utility and its remarkable accuracy in identifying and localizing lung pathologies.

\subsection{Fairness Evaluation}
In addressing the long-tailed problem, the tail class suffers from an extremely small sample size. This small sample size makes the tail class highly vulnerable because even slight demographic distribution biases can result in significant proportional biases. Consequently, it is essential to evaluate how a method's performance may vary across different demographic attributes.

The experiment was conducted on the test set of Pruned MIMIC-CXR-LT, excluding the `Calcification of the Aorta' and `Tortuous Aorta' categories, as certain demographic groups lack positive labels for these conditions. We assess fairness with two demographic attributes: race and gender. Race is categorized into five groups: White, Black, Hispanic, Asian, and Other. Gender is differentiated into two groups: male and female. In Table~\ref{table:fairness_version2}, higher EO is desirable, as it indicates uniform FNR across different demographic groups. In the `All’ result, in terms of race, data augmentation demonstrates the best EO performance, while both the ML-Decoder and ensemble methods failed to improve EO. However, for gender, although data augmentation boosts performance, it doesn't fully offset the drop caused by using ML-Decoder. ML-Decoder continues to negatively impact EO, while the ensemble method shows mixed results, performs well on tail classes but poorly on head classes. Thus, in contexts where FNR is critical, such as disease screening, data augmentation emerges as a promising approach. Prior studies have also confirmed the effectiveness of data augmentation techniques in improving model fairness.~\cite{cite:dataAug4fairness1,cite:dataAug4fairness2}.


When examining the `Head' and `Tail' separately, as shown in Table~\ref{table:fairness_version2}, first, the `Head' generally demonstrates better fairness performance. This is likely due to the smaller sample size of the `Tail' classes, which leads to increased bias. Second, our method positively impacts fairness for the `Tail', indicating its effectiveness for long-tailed classification where unbiased evaluation of the tail classes is crucial.

\begin{table*}[ht]
  \centering
  \begin{tabular}{|c|cccc|cc|}
  \hline
  \cellcolor{gray!25} &
    \multicolumn{4}{c|}{\cellcolor{gray!25}\textbf{Model}} &
    \multicolumn{2}{c|}{\cellcolor{gray!25}\textbf{Demographics}} \\ \cline{2-7} 
  \multirow{-2}{*}{\cellcolor{gray!25} \makecell{\textbf{Disease} \\ \textbf{Group}}} &
    \multicolumn{1}{c|}{\textbf{ConvNeXt}} &
    \multicolumn{1}{c|}{\textbf{ML-Dec.}} &
    \multicolumn{1}{c|}{\textbf{Data Aug.}} &
    \textbf{Ensemble} &
    \multicolumn{1}{c|}{\textbf{Race}} &
    \textbf{Gender} \\ \hline
  \multirow{4}{*}{All} & 
    \multicolumn{1}{c|}{$\checkmark$} &
    \multicolumn{1}{c|}{$\times$} &
    \multicolumn{1}{c|}{$\times$} &
    $\times$ &
    \multicolumn{1}{c|}{0.449 $\pm$ 0.243} &
    \textbf{0.715 $\pm$ 0.295} \\ \cline{2-7} 
   &  
    \multicolumn{1}{c|}{$\checkmark$} &
    \multicolumn{1}{c|}{$\checkmark$} &
    \multicolumn{1}{c|}{$\times$} &
    $\times$ &
    \multicolumn{1}{c|}{0.461 $\pm$ 0.261} &
    0.615 $\pm$ 0.320 \\ \cline{2-7} 
   &  
    \multicolumn{1}{c|}{$\checkmark$} &
    \multicolumn{1}{c|}{$\checkmark$} &
    \multicolumn{1}{c|}{$\checkmark$} &
    $\times$ &
    \multicolumn{1}{c|}{\textbf{0.495 $\pm$ 0.269}} &
    0.683 $\pm$ 0.302 \\ \cline{2-7} 
   &  
    \multicolumn{1}{c|}{$\checkmark$} &
    \multicolumn{1}{c|}{$\checkmark$} &
    \multicolumn{1}{c|}{$\checkmark$} &
    $\checkmark$ &
    \multicolumn{1}{c|}{0.479 $\pm$ 0.287} &
    0.678 $\pm$ 0.288 \\ \noalign{\hrule height 2pt}
  \multirow{4}{*}{Tail} &  
    \multicolumn{1}{c|}{$\checkmark$} &
    \multicolumn{1}{c|}{$\times$} &
    \multicolumn{1}{c|}{$\times$} &
    $\times$ &
    \multicolumn{1}{c|}{0.332 $\pm$ 0.265} &
    0.588 $\pm$ 0.335 \\ \cline{2-7} 
   &  
    \multicolumn{1}{c|}{$\checkmark$} &
    \multicolumn{1}{c|}{$\checkmark$} &
    \multicolumn{1}{c|}{$\times$} &
    $\times$ &
    \multicolumn{1}{c|}{0.311 $\pm$ 0.258} &
    0.478 $\pm$ 0.351 \\ \cline{2-7} 
   &  
    \multicolumn{1}{c|}{$\checkmark$} &
    \multicolumn{1}{c|}{$\checkmark$} &
    \multicolumn{1}{c|}{$\checkmark$} &
    $\times$ &
    \multicolumn{1}{c|}{\textbf{0.367 $\pm$ 0.291}} &
    0.544 $\pm$ 0.337 \\ \cline{2-7} 
   &  
    \multicolumn{1}{c|}{$\checkmark$} &
    \multicolumn{1}{c|}{$\checkmark$} &
    \multicolumn{1}{c|}{$\checkmark$} &
    $\checkmark$ &
    \multicolumn{1}{c|}{0.343 $\pm$ 0.321} &
    \textbf{0.607 $\pm$ 0.352} \\ \noalign{\hrule height 2pt}
  \multirow{4}{*}{Head} &  
    \multicolumn{1}{c|}{$\checkmark$} &
    \multicolumn{1}{c|}{$\times$} &
    \multicolumn{1}{c|}{$\times$} &
    $\times$ &
    \multicolumn{1}{c|}{0.599 $\pm$ 0.121} &
    \textbf{0.843 $\pm$ 0.140} \\ \cline{2-7} 
   &  
    \multicolumn{1}{c|}{$\checkmark$} &
    \multicolumn{1}{c|}{$\checkmark$} &
    \multicolumn{1}{c|}{$\times$} &
    $\times$ &
    \multicolumn{1}{c|}{0.630 $\pm$ 0.120} &
    0.756 $\pm$ 0.179 \\ \cline{2-7} 
   &  
    \multicolumn{1}{c|}{$\checkmark$} &
    \multicolumn{1}{c|}{$\checkmark$} &
    \multicolumn{1}{c|}{$\checkmark$} &
    $\times$ &
    \multicolumn{1}{c|}{\textbf{0.644 $\pm$ 0.132}} &
    0.818 $\pm$ 0.145 \\ \cline{2-7} 
   &  
    \multicolumn{1}{c|}{$\checkmark$} &
    \multicolumn{1}{c|}{$\checkmark$} &
    \multicolumn{1}{c|}{$\checkmark$} &
    $\checkmark$ &
    \multicolumn{1}{c|}{0.637 $\pm$ 0.115} &
    0.746 $\pm$ 0.153 \\ \hline
  \end{tabular}
  \caption{Evaluation of fairness using EO metric across disease groups among demographic attributes: race and gender.}
  \label{table:fairness_version2}
\end{table*}

\subsection{Comparison with Previous Approaches}
In this section, we explore various approaches used for dealing with dataset imbalance and multi-label classification. The subsequent article will briefly introduce these methods, outline the experimental configurations and provide insights into the potential reasons for their lack of success.

\paragraph{Feature decoupling cRT}
In the field of single-label long-tail datasets, the feature decoupling cRT method, is considered highly effective in previous research~\cite{cite:feature_decoupling}. However, our dataset's multi-label characteristic, featuring overlapping labels, rendered traditional re-sampling for balanced data impractical. Thus, we implemented a modified re-sampling approach, ensuring each class had at least 0.7 times the occurrence of the smallest class. We generated five distinct datasets with different random seeds, which were then used in the second stage of the feature decoupling cRT method to fine-tune the model.

In our investigation, we discovered that feature decoupling cRT did not improve performance, likely due to two main factors. First, the re-sampled dataset remained unbalanced, with the largest class being five times bigger than the smallest class. Second, prior research~\cite{cite:singleLabel_benchmark} indicates that although feature decoupling cRT improves balanced accuracy (the weighted mean of accuracy) on imbalanced test sets, it slightly decreases the mF1 score. Given that mAP is a similar metric, which considers both precision and recall, this may explain its sub-optimal performance.

\paragraph{Weighted loss}
We experiment two variants of weighted loss to manage the class imbalance. The first variant was the weighted binary cross entropy loss, where we determined the class weights by dividing the count of negative samples by the count of positive samples for each class. The second variant is the focal loss~\cite{cite:focal_loss}, where the alpha parameter was inversely related to the class performance, and we set the gamma parameter to a value of two as the original paper recommended. These weighted loss functions aimed to adjust the model's focus on classes with fewer samples and to mitigate the dominance of more prevalent classes.

The ineffectiveness of weighted loss arises from several key issues. First, by assigning higher weights to minority classes, the performance of previously well-performing classes is adversely affected. In particular, because the tail classes are so small, they require a disproportionately large weight, which makes the model overly sensitive to these classes. This sensitiViTy often fails to result in improvements for the weaker classes. Consequently, this leads to a decrease in mAP. Additionally, in multilabel datasets, where classes are not independent, heavily weighting certain labels can disrupt the learning of complex inter-dependencies between classes, further degrading overall model performance.

\paragraph{Random oversampling}
Random oversampling is a technique used to address class imbalance in datasets by increasing the number of instances in under-represented classes. This is achieved by randomly replicating instances from these classes until a more balanced distribution is reached. In our study, each class with fewer instances than a predetermined threshold was augmented by randomly duplicating images until the class size met this threshold. It's worth noting that in the multi-label dataset, when we oversample the minority classes, there's a possibility of oversampling the majority classes concurrently.  Despite the theoretical benefits of random oversampling in addressing class imbalance, our study observed performance degradation. This outcome can be attributed to overfitting on minority classes, as the repeated use of the same images in oversampling may reduce the model's generalization ability.

\paragraph{Self-supervised learning}
Self-Supervised Learning (SSL) is a machine learning approach that trains models on datasets without requiring labeled data. In image classification tasks, particularly for long-tailed datasets, an SSL pre-trained backbone is believed to help create a more balanced feature space, mitigating biases caused by uneven class representations. SimCLR~\cite{cite:SimCLR}, a prominent SSL method, enhances representation learning by maximizing agreement between various augmented versions of the same data sample. In our research, we employed SimCLR to pre-train a backbone for feature extraction from CXRs without using labels. This pre-trained backbone was then adapted for a multi-label classification task.

However, the SimCLR pre-trained backbone resulted in a decline in classification performance. This can be attributed to several factors. Firstly, the data augmentation techniques used in SimCLR might distort important features in medical images and the subtle and complex patterns in CXRs might not be effectively captured. Moreover, the severe class imbalance in long-tailed datasets may not be well-handled when fine tuning. Finally, the presence of watermarks on some CXRs can taint the features learned by the unsupervised method, as it lacks label information to distinguish relevant features from artifacts.

\begin{table}[ht]
\centering
\begin{tabular}{|l|l|l|}
\hline
\rowcolor{gray!25} \textbf{Method} & \textbf{Val.} & \textbf{Test} \\
\hline
Baseline & 0.376 & 0.372 \\
\hline
Ensemble (LTCXNet) & \textbf{0.384} & \textbf{0.377} \\
\hline
Feature Decoupling cRT~\cite{cite:feature_decoupling} & 0.356 & 0.347 \\
\hline
Focal Loss~\cite{cite:focal_loss} & 0.346 & 0.338 \\
\hline
Weighted Binary Cross Entropy Loss & 0.323 & 0.320 \\
\hline
ROS Oversampling~\cite{cite:multi-label_review} & 0.342 & 0.331 \\
\hline
SimCLR~\cite{cite:SimCLR} & 0.330 & 0.326 \\
\hline
\end{tabular}
\caption{Testing and validation mAP of various methods implemented with the baseline (ConvNeXt + ML-Decoder + Data Augmentation).}
\label{table:failed_methods_table}
\end{table}

\subsection{Clinical Feasibility}
Our method, LTCXNet, has a computational cost of 35 GFLOPs and can infer a single CXR within a second on a cost-effective GTX 1080, demonstrating its feasibility for clinical use. This efficient performance indicates that LTCXNet can be seamlessly integrated into clinical workflows, providing timely and accurate diagnostic support. 

\section{Conclusion}
We introduce LTCXNet to address the challenges of long-tailed, multi-label classification tasks. Through the evaluation of various methodologies on the Pruned MIMIC-CXR-LT dataset, we have identified a configuration that achieves optimal performance and significantly enhances the outcomes for the tail classes. Beyond performance, we also examine the impact of these methods on the model's fairness and the clinical practicability of LTCXNet. This research aims to advance the field of long-tailed, multi-label classification in medical imaging, contributing to more accurate and fair diagnostic tools.

{\small
\bibliographystyle{ieee_fullname}
\bibliography{egbib}
}

\end{document}